# Balanced Symmetric Cross Entropy for Large Scale Imbalanced and Noisy Data

Feifei Huang
Xiamen University
huangfeifei@aidigger.com

Jie Li
University of Science and Technology of China
lijie@aidigger.com

Xuelin Zhu
Southeast University
zhuxuelin23@gmail.com

## Abstract

*Deep convolution neural network has attracted many attentions in large-scale visual classification task, and achieves significant performance improvement compared to traditional visual analysis methods. In this paper, we explore many kinds of deep convolution neural network architectures for large-scale product recognition task, which is heavily class-imbalanced and noisy labeled data, making it more challenged. Extensive experiments show that PNASNet achieves best performance among a variety of convolutional architectures. Together with ensemble technology and negative learning loss for noisy labeled data, we further improve the model's performance on online test data. Finally, our proposed method achieves 0.1515 mean top-1 error on online test data.*

1. Introduction

Benefit from ImageNet data, deep neural network has received significant developments, and has been proven to be an effective mothed for copying with a variety of visual tasks. During past decade, many kinds of convolutional architectures have been framed for automatic visual feature extraction, and achieve significant breakthrough in large-scale image classification task. Moreover, as the backbone of the neural network, their pretrained models on ImageNet have supported a large number of downstream tasks, such as object detection [13, 14], semantic segmentation [15, 16], pose estimation [17, 18], image caption [19] and so on. There is no doubt that deep learning has become a dominating technology in visual task and attracted increasing enthusiasm of computer vision researchers.

Unlike common image classification task, large-scale product recognition task is more challenged since there are so many highly similar products in product image dataset, which requires neural network to learn more detailed and distinct visual features among different products, making the task quite intractable. In this work, we explore many popular deep convolutional architectures for large-scale product recognition, including ResNet [1], SENet [2], ResNeSt [3] and PNASNet [4]. Besides, we also introduce negative learning loss named symmetric cross entropy (SCE) in order to alleviate the noise in data, which boosts cross entropy symmetrically with a noise robust counterpart reverse cross entropy. In addition, test-time augmentation for test data is applied for notably boost model's performance on online test data.

2. Related work

As the pioneer in deep learning, AlexNet [5] achieves electrifying performance in ILSVRC-2012 competition, and motivates increasing research enthusiasm in convolutional architectures, making deep learning the dominant technology in the field of computer vision. With this trend, VGG-Net [6] proposes a modular network design strategy that simplifies the workflow of network design and transfer learning for downstream applications by stacking the same type of blocks repeatedly. ResNet [1] introduces an identity skip connection which mitigates the dilemma of vanishing gradient in deep neural network and enables network learning deeper feature representations. ResNeSt, one of the variants of ResNet, generalizes the channel-wise attention into feature-map group representation, and improves downstream tasks performance by stacking Split-Attention blocks. Based on neural architecture search (NAS [7]) method, PNASNet uses a sequential model-based optimization (SMBO) strategy and searches for structures in order of increasing complexity, while simultaneously learning a surrogate model to guide the search through structure space.

Modern deep neural networks commonly contain millions of trainable parameters that require large scale datasets with clean labeled data to learn, such as ImageNet and COCO. However, it is expensive to collect such high-quality labeled data. So far, a great number of methods have been proposed to solve the overfitting on noisy labeled data without any extra clean labeled data. Li *et al.* proposed meta learning based noise-tolerant (MLNT), which applied the MAML method in meta learning and mean teacher method in self-ensembling technology [9]. Wang et al. applied negative learning loss (SCE) called symmetric cross entropy to overcome the overfitting on noisy data [10]. In our experiment, SCE can significantly improve mean top-1 error on Aliproducts dataset.

## 3. Solution for challenge

The large-scale product recognition challenge refers to multi-dimensional challenges, including heavily class-imbalanced data, noisy labeled data, a large number of classes (~50K) and the trade off between coarse-grained classification and fine-grained classification.

To overcome aforementioned challenges, we first introduce the balanced symmetric cross entropy, which has been proven to be an effective method for tackling noisy data. Cross entropy is well known as a kind of distance measurement between two probability distributions (as the Eq.1 shown):

$$CE = -\sum_{k=1}^{K} q(k|x) \log p(k|x) \quad (1)$$

where $q(k|s)$ is ground truth class distribution conditioned on sample $x$, $p(k|x)$ is the predicted class distribution by neural network on sample $x$. $k$ is the number of classes in dataset.

To solve the imbalanced class distribution problem in data, we introduce a balancing factor $w$ in cross entropy:

$$BCE = -\sum_{k=1}^{K} w(k) \cdot q(k|x) \log p(k|x) \quad (2)$$

and the factor $w$ is computed as follow:

$$w(k) = \frac{N}{K \cdot n(k)} \quad (3)$$

Where $N$ is the number of samples in training dataset, $N(k)$ is the number of samples of class $k$ [12].

However, cross entropy is a positive learning loss, which heavily relies on the correctness of annotations. When data is noisily labeled, $q(k|s)$ will be not able to represent the true class distribution, which will cause a dilemma that cross entropy force $p(k|x)$ to learn such kind of incorrect distribution. Instead, negative learning is proposed to learn that which classes the input $x$ does not belong to. As an effective negative loss, the reverse cross entropy is defined as:

$$RCE = -\sum_{k=1}^{K} p(k|x) \log q(k|x) \quad (4)$$

Then the balanced symmetric cross entropy can be defined as:

$$BSCE = \alpha \cdot BCE + \beta \cdot RCE \quad (5)$$

Where $\alpha$ and $\beta$ are hyper-parameters. The RCE term is noise tolerant while the BCE term is useful for achieving good convergence. Such kind of weighted combination allows more effective and robust learning for deep neural networks [10].

With the help of the balanced symmetric cross entropy, deep neural networks achieve more robust performance on Aliproducts validation dataset. Moreover, we also employ center crop test time augmentation to further boost the performance of single model. Finally, we ensemble different models together using vote strategy, thus achieving competitive online score. More details are introduced in section 4.

## 4. Experiments

We explore many kinds of convolutional neural network architectures, including ResNet, SENet, ResNeSt and PNASNet. Before training on the Aliproducts dataset, all these networks are initialized with pretrained weights on the ImageNet. Stochastic gradient decent is used to further optimize networks' weights on Aliproducts dataset, and the initial learning rate is set as 0.01. During the training, we reduce the learning rate by one-tenth when the mean top-1 error on validation data plateaus. The hyper-parameters $\alpha$ and $\beta$ in balanced symmetric cross entropy are empirically set as 0.4 and 0.7 respectively.

### 4.1. Backbone experiments

We first conduct extensive experiments with variant backbone on Aliproducts dataset. The mean top-1 errors of different backbones on validation dataset are shown in Table 1. The results indicate that PNASNet performs much better than Resnet152 and SENet154, and slightly better than ResNeSt, which is proposed recently and lead to state of the art in various downstream applications.

Table 1: Mean top-1 errors of different backbones on Aliproducts validation dataset.

| Backbone | ResNet152 | SENet154 | ResNeSt 269 | PNAS Net |
|---|---|---|---|---|
| Mean top-1 error | 0.2169 | 0.2248 | 0.1584 | **0.1530** |

### 4.2. Loss experiments

In this section, we compare the performance of PNASNet with different loss functions, including cross entropy loss (ce), balanced cross entropy loss (bce), symmetric cross entropy loss (sce) and balanced symmetric cross entropy loss (bsce). The results are shown in Table 2. According to the table 2, the performance of PNASNet is improved to 0.1407 with the help of the bsce loss.

Table 2: Mean top-1 error of different loss on PNASNET on Aliproducts validation dataset.

| Loss | ce | bce | sce | bsce |
|---|---|---|---|---|
| Mean top-1 error | 0.1530 | 0.1430 | 0.1492 | **0.1407** |

### 4.3. Center crop experiments

During the experiment, we find that the center crop on product images can notably boost the performance of models. In our center crop experiment, we first resize image

to different sizes, including 384x384, 412x412, 424x424, 436x436 and 464x464, then crop the resized image to the PNASNet input size (331x331). Finally, we ensemble these scales by averaging the features extracted by PNASNet to augment the test images, which is known as test-time augmentation (tta). As the Table 3 shows, tta can achieve better performance on validation dataset.

Table 3: Mean top-1 error of different crop size and tta results of PNASNet on Aliproducts validation dataset.

| Size (w=h) | 384 | 412 | 424 | 436 | 464 | tta |
|---|---|---|---|---|---|---|
| Mean top-1 error | 0.1407 | 0.1386 | 0.1384 | 0.1383 | 0.1425 | **0.1354** |

4.4. Ensemble experiment

Finally, we ensemble all models by top-1 voting strategy, including SENet, PNASNet and ResNeSt with bsce loss, thus achieving 0.1515 mean top-1 error on online test data.

Table 4: Mean top-1 error of the model ensemble on Aliproducts online test dataset.

| Model | ensemble |
|---|---|
| **Mean top-1 error** | **0.1515** |

5. Conclusion

Main contribution of this work is to show how we improve models' performance for large-scale product image recognition by using balanced symmetric cross entropy in different convolution neural network architectures. Integration technology is also important for large-scale product recognition task. There are many possible directions for future work, such as exploring better noisy data processing strategy, introducing tail-augmentation methods for improving the variance of tail classes.